\documentclass[letterpaper, 10 pt, conference]{ieeeconf}  % Comment this line out if you need a4paper

\usepackage{amsthm}

\IEEEoverridecommandlockouts                              % This command is only needed if 
\pdfoutput=1 
\usepackage{commath}
\usepackage{textcomp}
\let\labelindent\relax
\usepackage{enumitem}
\usepackage[ruled,linesnumbered,resetcount,vlined]{algorithm2e}
\usepackage{graphicx} % for pdf, bitmapped graphics files
\usepackage{caption}
\usepackage{subcaption}
\usepackage{multicol}
\usepackage{amsmath}
\usepackage{amssymb}
\usepackage{amsthm}

\usepackage[style=ieee,sorting=none,backend=biber,giveninits=true]{biblatex}
\usepackage[bookmarks=true]{hyperref}
\usepackage[capitalize]{cleveref}
\usepackage{float}
\usepackage{multirow}
\usepackage{tabularx}
\usepackage[usenames, dvipsnames, table]{xcolor}

\colorlet{mylinkcolor}{BrickRed}
\colorlet{mycitecolor}{Green}
\colorlet{myurlcolor}{NavyBlue}

\hypersetup{
  linkcolor  = mylinkcolor,
  citecolor  = mycitecolor,
  urlcolor   = myurlcolor,
  colorlinks = true,
}

\theoremstyle{definition}

\usepackage{subcaption}
\begin{document}

% \title{\LARGE \bf
% Towards Super-Nominal Payload Manipulation:\\A Multi-Skill Approach with Diffusion-Based Motion Planning
% }

\title{\LARGE \bf
Towards Super-Nominal Payload Handling:\\Inverse Dynamics Analysis for Multi-Skill Robotic Manipulation
}

\author{Anuj Pasricha$^{*}$ and Alessandro Roncone%
\thanks{$*$ Corresponding author.}%
\thanks{The authors are with the Department of Computer Science, University of Colorado Boulder, 1111 Engineering Drive, Boulder, CO USA. This work was supported by the Office of Naval Research under Grant N00014-22-1-2482. {\tt\small firstname.lastname@colorado.edu}}%
}

\maketitle
\thispagestyle{empty}
\pagestyle{empty}

\begin{abstract}
Motion planning for articulated robots has traditionally been governed by algorithms that operate within manufacturer-defined payload limits.
Our empirical analysis of the Franka Emika Panda robot demonstrates that this approach unnecessarily restricts the robot's dynamically-reachable task space.
% We show that these robots can transport heavier payloads, even exceeding twice their nominal capacity.
These results establish an expanded operational envelope for such robots, showing that they can handle payloads of more than twice their rated capacity.
Additionally, our preliminary findings indicate that integrating non-prehensile motion primitives with grasping-based manipulation has the potential to further increase the success rates of manipulation tasks involving payloads exceeding nominal limits.
% Our analysis demonstrates the critical need for integrating a more nuanced consideration of payload dynamics into motion planning algorithms.
% Our results showcase an expanded operational envelope, highlighting the importance of a more nuanced consideration of payload dynamics in motion planning algorithms.
\end{abstract}

\section{Motivation}\label{sec:introduction}

The design of robotic systems fundamentally shapes their manipulation capabilities and therefore, directly influences their potential to provide increased productivity gains across various industries \cite{van2023investigating}.
% The potential of robotic manipulation to provide increased productivity gains across various industries is well-documented \cite{billard2019trends}.
For articulated robots, design choices such as link dimensions and motor specifications inform torque limitations, which in turn induce a maximum payload capacity.
% and payload limits are set conservatively to ensure absolute safety, aiming for 100\% motion success even in worst-case scenarios (eg: around singular configurations).
These payload limits are often determined by the manufacturer in a manner that is opaque to the end user---they are typically calculated based on the worst-case scenario and are assigned a single value across the robot's entire configuration space, which results in a significant under-utilization of robotic capabilities.
% This results in low nominal payload capacities, which significantly limit the operational space of these robots.
% Moreover, the pick-and-place-centric perspective that is dominant in robotic manipulation literature and industry applications compounds this issue \cite{calli2015benchmarking}.
Moreover, the predominant focus on pick-and-place manipulation in both robotics research and industry exacerbates this under-utilization, neglecting the diverse manipulation skills that could exploit the robot's full capabilities across its configuration space \cite{calli2015benchmarking}.

In this work, we present an analysis showing that robot manipulators can safely handle payloads exceeding their factory limits, potentially \textsl{up to twice their nominal capacity}.
% This finding is especially crucial for collaborative robots, which are typically lightweight and designed with strict safety margins to ensure absolute safety, even in worst-case scenarios.
Additionally, we demonstrate that supplementing grasping-based manipulation with a multi-skill approach, specifically by incorporating non-prehensile motion primitives such as pushing, can further increase the chances of successful manipulation, underscoring the need for complementary skills in motion planning \cite{stuber2020let, pasricha2022pokerrt}. This work contrasts with the limited existing research on super-nominal payload manipulation, which primarily focuses on point-to-point motion planning for non-redundant robot arms \cite{wang2001payload, gallant2018extending}.

% \input{2.background}
% \section{Methods}\label{sec:methods}

\begin{figure*}
\centering
\includegraphics[width=2\columnwidth,height=0.897\columnwidth]{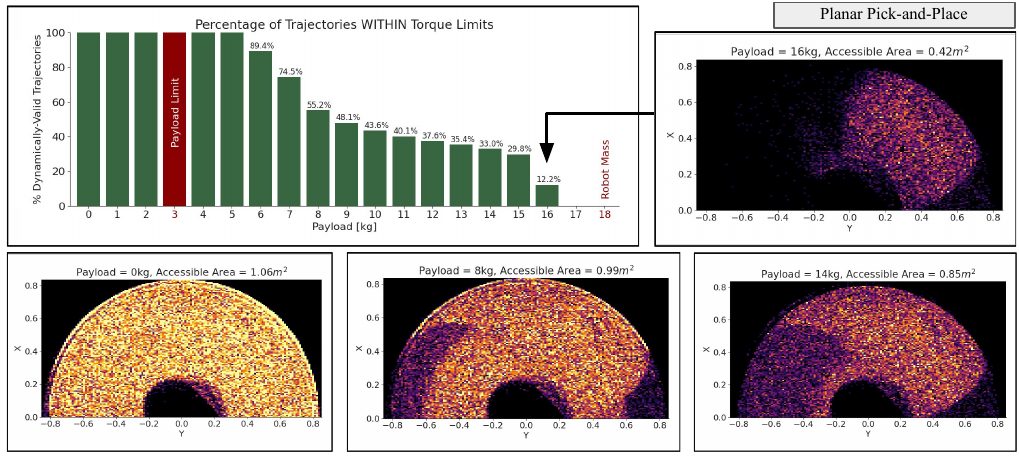}
\caption{Despite a decrease in dynamically-valid trajectories for increasing payloads, the robot maintains significant functionality at 2.7 times the nominal payload of 3 $kg$. It retains a reachable area of 0.99 $m^2$ at 8 $kg$, only a slight decrease from 1.06 $m^2$ at 3 $kg$.
}
\label{fig:pnp}
\end{figure*}

\begin{figure*}
\centering
\includegraphics[width=2\columnwidth,height=0.897\columnwidth]{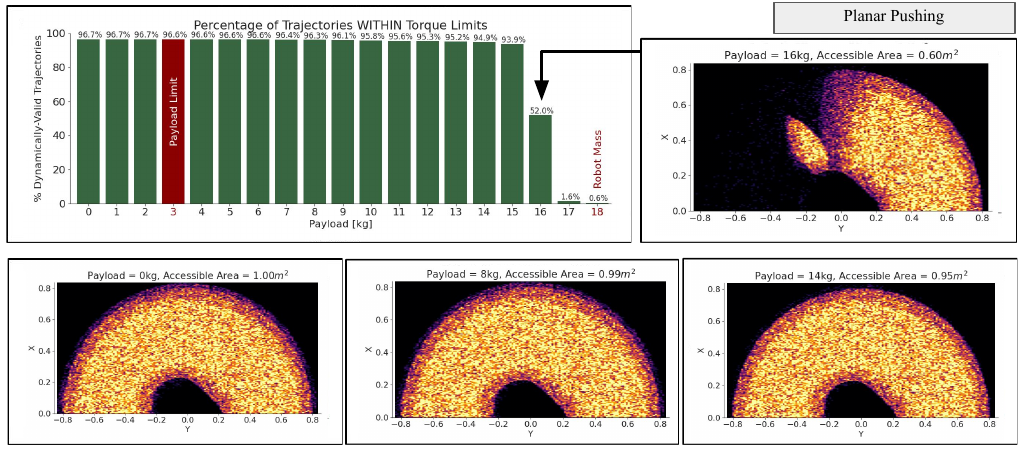}
\caption{When pushing, the robot maintains high percentages of dynamically-valid trajectories up to 15 $kg$, with a significant drop beyond (top left). Even at 2.7 times the nominal payload of 3 $kg$, the robot retains substantial functionality, maintaining a reachable area of 0.99 $m^2$ at 8 $kg$, only slightly reduced from 1.00 $m^2$ at 0 $kg$.}
\label{fig:push}
\end{figure*}

\section{Payload-Aware Torque Constraints}\label{sec:static}

Torque constraints are applied to a robot's trajectory as defined by $\vec{x} = \{q, \dot{q}, \ddot{q}\}$, where $q$ is the target position, $\dot{q}$ is the commanded joint velocity, and $ \ddot{q} $ is the required acceleration based on the robot's current velocity.
The required torque $\tau_i$ for state $\vec{x}_i$ is calculated using the Recursive Newton-Euler Algorithm (RNEA) using forward and backward recursion over the robot's links.
In the forward recursion step, RNEA calculates the linear and angular velocities and accelerations for each link, starting from the base link. Using these values, backward recursion begins at the end-effector link, calculating the forces induced by the accelerations on each link and determining the effective torque at the associated joints.
In our analysis, we treat the payload as an additional link rigidly attached to the end-effector. The torque constraint is satisfied if none of the calculated torques $\tau_i$ for any joint $i$ exceed the maximum torque limit of that joint $\tau_{i, max}$.

% \section{Evaluation}\label{sec:evaluation}

% \begin{itemize}
%     \item Static and trajectory payload analysis
%     \item Skill-based analyses
% \end{itemize}

% => 

\section{Data Collection}

We use cuRobo to generate 100,000 trajectories for both pick-and-place and pushing using the 7DoF Franka Emika Panda robot arm \cite{sundaralingam2023curobo}. For each trajectory, a payload is rigidly attached to the robot's end-effector in simulation, and every waypoint ($q$, $\dot{q}$, $\ddot{q}$) for a given trajectory is verified by RNEA to ensure that the robot's torque limits are satisfied. We emphasize that this analysis procedure is robot-agnostic.

\section{Pick-and-Place Trajectory Analysis}

Pick-and-place trajectories are generated for random end-effector poses on a fixed planar surface, with the end-effector oriented for top-down grasping. Analysis for these trajectories under varying payload conditions is presented in \cref{fig:pnp}. The bar chart (top left) illustrates the percentage of dynamically-valid trajectories within torque limits for payloads ranging from 0-18 $kg$, demonstrating a noticeable, but expected, decrease in performance as the payload increases. The accessible workspace areas for specific payloads are depicted as follows: 0 $kg$ (bottom left, 1.06 $m^2$), 8 $kg$ (bottom center, 0.99 $m^2$), 14 $kg$ (bottom right, 0.85 $m^2$), and 16 $kg$ (top right, 0.42 $m^2$). These workspace plots indicate that the robot remains capable despite reduced reachability with higher payloads. Notably, the robot is capable of nearly doubling the nominal payload capacity (highlighted in red) while maintaining significant functionality in the super-nominal payload regime.

\section{Push Trajectory Analysis}

Push trajectories consist of planar motions that span 10$cm$ in distance, where the direction of motion is chosen in increments of 45$^{\circ}$.
These motions are executed at speeds chosen to ensure operation within the quasi-static realm.
The analysis of planar pushing performance, illustrated in \cref{fig:push}, reveals that the robot maintains high percentages of dynamically-valid trajectories within torque limits up to a payload of 15 $kg$, with a significant decline beyond this threshold.
Even at 2.7 times the nominal payload capacity of 3 $kg$, the robot retains substantial functionality, maintaining a reachable area of 0.99 $m^2$ at 8 $kg$, which is only slightly reduced from 1.00 $m^2$ at 0 $kg$.
% This highlights the robot's robust performance under increased load conditions.
With pushing demonstrating comparatively higher success rates for high payloads, these results indicate that pushing and pick-and-place operations are complementary in nature and are both required to maximize a given embodiment's manipulation capability.
% This complementary nature suggests that pushing strategies can enhance the overall efficiency and reliability of the robot's performance, particularly in handling increased payloads.

% Emphasize that still able to nearly double the nominal payload capacity with pick and place

% pushing and pick and place are complementary
% higher success rates for pushing for the most part
\section{Conclusion}\label{sec:future-work}

Our initial analysis demonstrates that a configuration-specific understanding of the task space, combined with non-prehensile motion primitives, enables the manipulation of heavier payloads. While pick-and-place motions show a decrease in performance with higher payloads, pushing motions generally exhibit high success rates, emphasizing the need for multiple complementary skills to expand the operational space of existing robot embodiments. We are currently incorporating the analysis presented in this extended abstract into a motion planning framework to enable the successful manipulation of super-nominal payloads \cite{orthey2023sampling, fishman2023motion, pasricha2024virtues}.

% \clearpage
\AtNextBibliography{\footnotesize}
\printbibliography

\end{document}